\documentclass[11pt,a4paper]{article}
\usepackage[hyperref]{acl2021}
\usepackage{times}
\usepackage{latexsym}

\usepackage{graphicx}
\usepackage{url}
\usepackage{amsmath}
\usepackage{breqn}
\usepackage{makecell}
\usepackage{tabularx}
\usepackage{booktabs}
\usepackage{multirow}
\usepackage[colorinlistoftodos]{todonotes}
\usepackage{textcomp}
\usepackage{bm}

\usepackage{hyperref}
\usepackage{amsmath}
\usepackage{microtype}
  \definecolor{DarkGreen}{rgb}{0.2,0.5,0.2}

\newcommand*{\affaddr}[1]{#1}
\newcommand*{\affmark}[1][*]{\textsuperscript{#1}}

\usepackage{microtype}

\newcommand{\IntroductionMotivatingExample}{
\begin{table}[h!]
\centering
\small
\begin{tabularx}{\columnwidth}{|X|}
\hline
\textbf{Question:} i found a puppy that is less then six weeks old an no mother around what should i feed it? 
 \\ \hline
\textbf{Context:} it a pit puppy i think
 \\ \hline
\textbf{Answer 1:} \textcolor{red}{Go to a vet} and \textcolor{blue}{get some and a small feeding bottle.} 
\\ \hline
\textbf{Answer 2:} \textcolor{blue}{get a baby bottle warm milk} best thing is to \textcolor{red}{call a pet shop}  \\ \hline
\textbf{Answer 3:} it needs a certain type of milk, \textcolor{purple}{ dont feed it cows milk}
\\ \hline
\textbf{Answer 4:} \textcolor{red}{call a vet and ask them.} if you cannot do that then give them alot of water and a little balony a day, \textcolor{DarkGreen}{than go into dog food...}
\\ \hline
\textbf{Summary Bullet Points:} \\ \hline
\textbf{1.} \textcolor{red}{call the vet} and tell them how old you think it is and what should you feed it... \\
\textbf{2.} the first thing you want to do if you plan to keep it is \textcolor{red}{go to petsamrt or pet co and ask anyone that specializes on dogs} and \textcolor{blue}{get the pup a baby bottle} and feed it milk but \textcolor{purple}{ not cow milk} try powder milk with water. \\
\textbf{3.}  Try and find something soft to eat (as in a soft \textcolor{DarkGreen}{ dog food).} \\
\textbf{4.} if it is not yet walking, \textcolor{blue}{then get a bottle} \\
\hline
\end{tabularx}
\caption{An example bullet-point summary from our answer summarization dataset, illustrating the multiple viewpoints present in the summaries created through our pipeline, and a subset of the 14 user answers to which the target summary can be aligned.}
\label{tab:example}
\end{table}
}

\newcommand{\DatasetComparison}{
 \begin{table}[t]
\resizebox{\columnwidth}{!}{\begin{tabular}{c c c c c}
\Xhline{2\arrayrulewidth}
             \textbf{Dataset} & \textbf{\% Novel unigrams} &  \textbf{Oracle Extractive}  &  \textbf{Length}    \\  \hline
            AnswerSumm (ours) & 32.2 &  40.02/11.16/33.70 & 67  \\ 
            XSUM & 35.8 & 29.79/8.81/22.65 & 23 \\
            CNN & 16.8 & 50.38/28.55/46.58 & 46 \\
            DailyMail & 17.0 & 55.23/30.55/51.24 & 55 \\
            \Xhline{2\arrayrulewidth}
\end{tabular}}
\caption{Comparison between AnwerSumm and the XSum \cite{narayan-etal-2018-dont} and CNN-DailyMail \cite{nallapati-etal-2016-abstractive} datasets. Oracle Extractive and Length refer to the maximum ROUGE \cite{lin-2004-rouge} score achievable by an extractive model, and the average length of the summaries, respectively.}
\label{tab:statistics}
\end{table}
}

\newcommand{\BaselineResults}{
\begin{table}[t!]
\small
\resizebox{\columnwidth}{!}{\begin{tabular}{|l|c|}
\hline
    \textbf{Method}   & \textbf{ROUGE-1/2/L}   \\ \hline
LexRank	&	26.86/5.17/22.68	\\ \hline
TextRank	&	27.44/5.05/22.13	\\ \hline
BertSum \cite{liu2019text} &	30.01/5.76/24.83	\\ \hline
\end{tabular}}
\caption{ROUGE scores for baseline extractive models.}
\label{tab:baseline_results}
\end{table}
}

\newcommand{\MainResults}{
\begin{table}[t!]
\resizebox{\columnwidth}{!}{\begin{tabular}{|l|cc|}
\hline
    \textbf{Method}   & \textbf{ROUGE-1/2/L}    & \textbf{NLI}       \\ \hline \hline
BART baseline &	\textbf{33.37}/\textbf{8.39}/29.41	&	48.13	\\
\hline
BART + RL (ROUGE) &	33.26/8.30/29.46	&	49.29	\\
BART + RL (NLI) &	33.05/8.36/29.23 &	\textbf{56.68}	\\
BART + RL (Semantic Area)	&	33.33/8.28/\textbf{29.60} &	51.14	\\
BART + RL (ALL) &	\textbf{33.54}/\textbf{8.41}/\textbf{29.65} &	51.18		\\
\hline
BART + Sent Relevance	&	32.95/8.33/29.38 &	52.31	\\
BART + Sent Relevance + RL (ALL) &	33.21/8.29/29.46	&	\textbf{56.99}	\\
\hline
\end{tabular}}
\caption{ROUGE and NLI scores for proposed models, with the two highest scores for each metric highlighted}
\label{tab:main_results}
\end{table}
}

\newcommand{\NLIMotivation}{
\begin{table}[t!]
\resizebox{\columnwidth}{!}{\begin{tabular}{|l|c|}
\hline
    \textbf{Method}   & \textbf{\% Correct}     \\ \hline \hline
BERT NLI (\cite{falke-etal-2019-ranking}) &	64.1\%	\\
FactCC \cite{kryscinskiFactCC2019}	&	70.0\%	\\
QAGs \citet{wang-etal-2020-asking}	&	72.1\%	\\ \hline
BART MNLI (sentence) &	71.9\%	\\
RoBERTa MNLI (sentence)	&	 89.8\%	\\ \hline
RoBERTa MNLI (article)	&	78.6\%	\\
RoBERTa MNLI (max article sentence)	&	85.0\%	\\

\hline
\end{tabular}}
\caption{Results from faithfulness ranking evaluation from \citet{falke-etal-2019-ranking}, showing the importance, both of the strength of the NLI model on downstream faithfulness performance,and the effect of input granularity on performance. Sentence and article in parentheses indicate the granularity of the source input to the NLI model; max sentence calculates the max score over all article sentences as the score of a given target sentence.}
\label{tab:nli_motivation}
\end{table}
}

\newcommand{\PipelineFigure}{
\begin{figure*}[t]
    \centering
    \includegraphics[width=2.0\columnwidth,height=5\textheight,keepaspectratio]{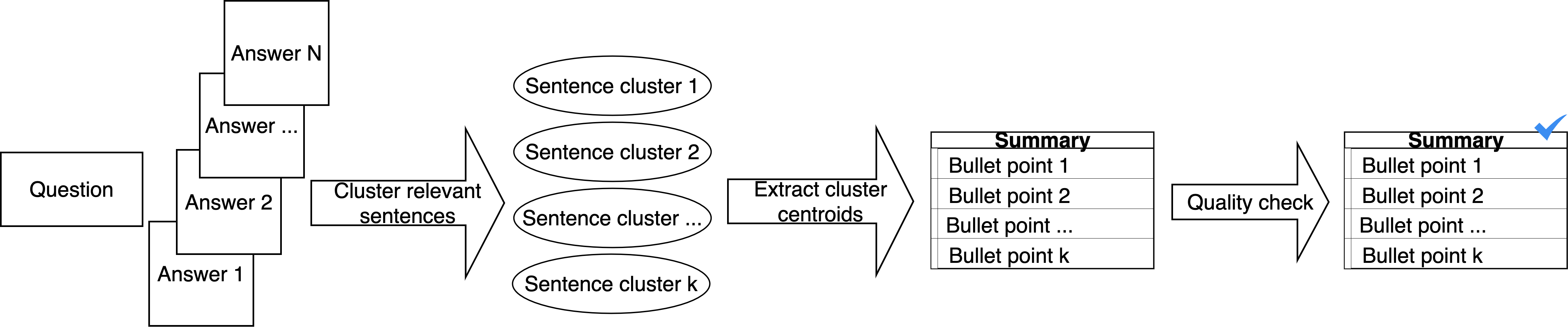}
    \caption{An illustration of our dataset pipeline. Given a question and answers, we cluster relevant sentences and remove the cluster centroid of non-singleton clusters from the input to use as bullet point summaries, filtering the example if it does not meet quality-control criteria.}
    \label{fig:pipeline}
\end{figure*} 
}

\newcommand{\HumanAnnotation}{
\begin{table}[t!]
\centering
\resizebox{\columnwidth}{!}{\begin{tabular}{c c c c  } \Xhline{2\arrayrulewidth}
             \textbf{Method} & \textbf{Multi-Perspective}& \textbf{Faithful}  \\ \Xhline{\arrayrulewidth}
             BART   & 4.45 & 3.72  \\ 
             BART + RL (ALL) & 4.57 & 4.13   \\ 
             BART + Sent Relevance + RL (ALL)   & 4.55 & 4.24  \\
             \Xhline{2\arrayrulewidth}
\end{tabular}}
\caption{Human evaluations of model outputs measuring the ability to capture multiple perspectives and faithfulness. Higher is better.}
\label{tab:human_evaluation}
\end{table}
}

\newcommand{\SpanPredictionExample}{
\begin{table}[t!]
\centering
\begin{tabularx}{\columnwidth}{X|X}
\multicolumn{2}{c}{\textbf{Question:} What is the secret to work/life balance?} \\ \hline
 \multirow{2}{*}{\textbf{Summary Sentences:}} &  \textbf{Associated Source}  \\
 &  \textbf{Sentences:} \\ \hline
You have to find the right balance between work and life.   & I mean you keep looking outside of work for happiness, and you want a balance, so why not be happy everywhere  \\  \hline
If you don't try something new, you'll never know what you're doing. & If what you're doing now isn't working, why not try something new \\  \hline
You have to make them both equal. & Only then will they matter equally \\  \hline
It's a great book, and you can get it at any book store. & It's absolutely possible, and in my sources is a book that you can get as cheap as \$1.62 \\  \hline
I think the trick is to go to work with the right attitude. & It seems to me that people just go to work with the wrong attitude actually \\
\hline
\end{tabularx}
\caption{An example of the predicted sentences from our span-based model with all rewards. On the left side are the generated summary sentences and on the right side are the sentences predicted to be relevant at the end of sentence timestep during generation. }
\label{tab:example_spans}
\end{table}
}

\newcommand{\ExampleSummaries}{
\begin{table}[t!]
\small
\centering
\begin{tabularx}{\columnwidth}{|X|}
\hline
\textbf{Question:} average 14 year old girls weight?
 \\ \hline
\textbf{Context:} im 14 years old, 145 pounds, 5'5'' and 1/2, need to know if i need to lose weight. my brother and his friend(who is also my friend) have told me i do, but i dont know. is this a normal weight?
 \\ \hline \hline
\textbf{Answer 1:} 
be comfortable in your own body, don't worry what others think of you! If you feel like you need to lose weight then exercise and make that choice for yourself.
\\ \hline
\textbf{Answer 2:} your bro and his friend are retards who cares what they say??? and if you think you weigh alot it probaly doesn't show  \\ \hline
\textbf{Answer 3:} First off, that is not very nice for your brother and his friends to be telling you to loose weight ...  It is actually more healthy to be about 10\% over what is normal for your age. Remember that muscle weights more than fat but it takes up less room than fat does. If you eat well and exercise daily you have nothing to worry about. \\ \hline
\textbf{Answer 4:} i'm 5'8" and wrestled at 126 lbs when i was a freshman and sophomore.
\\ \hline
\textbf{Answer 5:} Your BMI (Body Mass Index) is 24.1 Based on this number and your age you are not overweight, but are considered "at risk for overweight."
\\ \hline
\hline

\textbf{BART Baseline Summary:} You are not overweight if you are eating healthy and exercising. $\langle S \rangle$
 Your weight will change as you get older. $\langle S \rangle$ \textcolor{red}{If you are overweight, it is because you have too much muscle and not enough fat.} $\langle S \rangle$ \textcolor{red}{I'm not sure how old you are, but I'm assuming you are 14.} $\langle S \rangle$ \textcolor{red}{It is not a good idea to lose weight.}
 \\ \hline
\textbf{BART + RL (ALL) Summary:} If you feel like you need to lose weight then exercise and make that choice for yourself.  $\langle S \rangle$  You are not overweight, \textcolor{red}{and you are not at risk for overweight.}  $\langle S \rangle$ It is normal for a fourteen year old to be a little over weight, but not overweight.  $\langle S \rangle$ If you are a wrestler, you will know how much muscle you have.
 \\ \hline
\textbf{BART + Sent Relevance + RL (ALL) Summary:} If you feel like you need to lose weight then do so, but don't listen to your brother and his friend.  $\langle S \rangle$ You are not overweight, but you are at risk for being overweight.  $\langle S \rangle$ You should be comfortable with your weight.  $\langle S \rangle$ If you have muscle, you will be able to lose more weight than if you had fat.
 \\ \hline
\end{tabularx}
\caption{Example question and answers along with bullet-point answer summaries from three models. Possible hallucinations are shown in red.}
\label{tab:example_summaries_1}
\end{table}
}

\aclfinalcopy 

\title{Instructions for ACL-IJCNLP 2021 Proceedings}

\author{
 \textbf{Alexander R. Fabbri}\affmark[$\dagger$]    \quad \textbf{Xiaojian Wu}\affmark[$\ddagger$]
 \quad \textbf{Srini Iyer}\affmark[$\ddagger$]
 \quad \textbf{Mona Diab}\affmark[$\ddagger$] \\
\affaddr{\affmark[$\dagger$] Yale University} 
  \affaddr{\affmark[$\ddagger$] Facebook AI} \\
  \texttt{alexander.fabbri@yale.edu} \\
          \texttt{\{xiaojianwu,sviyer,mdiab\}@fb.com} 
}

\date{}
\title{Multi-Perspective Abstractive Answer Summarization}
\begin{document}
\maketitle

\begin{abstract}
Community Question Answering (CQA) forums such as Stack Overflow and Yahoo! Answers contain a rich resource of answers to a wide range of questions. Each question thread can receive a large number of answers with different perspectives. The goal of multi-perspective answer summarization is to produce a summary that includes all perspectives of the answer. A major obstacle for multi-perspective, abstractive answer summarization is the absence of a dataset to provide supervision for producing such summaries. This work introduces a novel dataset creation method to automatically create multi-perspective, bullet-point abstractive summaries from an existing CQA forum. Supervision provided by this dataset trains models to inherently produce multi-perspective summaries. Additionally, to train models to output more diverse, faithful answer summaries while retaining multiple perspectives, we propose a multi-reward optimization technique coupled with a sentence-relevance prediction multi-task loss. Our methods demonstrate improved coverage of perspectives and faithfulness as measured by automatic and human evaluations compared to a strong baseline.
\end{abstract}
\section{Introduction}\label{sec:introduction}
\IntroductionMotivatingExample
In a world of information overload and the ubiquity of discussion forums, there is a need for text summarization as a means of distilling relevant information into a concise form. The problem is even more pertinent for question answering within the context of Comunity Question Answering (CQA) forums, where a person poses a question and can get an abundance of answers to sift through. Ideally, an answer summary should cover the multiple perspectives found in the answers, where available. For example, in Table \ref{tab:example}, a person poses a question about finding a puppy and also provides context on the type of dog. We present a sample of the 14 answers to that question on Yahoo! Answers and an automatically-created summary consisting of bullet points covering the answers' main perspectives. We introduce a novel pipeline to build such a \textit{multi-perspective, bullet-point summarization dataset} and introduce models to generate faithful multi-perspective summaries.
\par
To date, most CQA forums have a notion of a 'best answer,' which is either manually chosen by the person who asked the question or by a moderator or obtained via community ratings. Work in this field typically makes use of this best answer as a proxy for summaries \cite{tomasoni-huang-2010-metadata,chan-etal-2012-community,pande2013summarizing,wang-etal-2014-query, song2017summarizing}. However, the best answer only presents one person's perspective and rarely captures the variety of perspectives discussed in the thread. Datasets such as WikiHowQA \citep{deng2020joint}, which consists of a question, a long answer, and an answer summary, focus on answer selection and the summarization of a single answer. While CQASumm \cite{chowdhury2019cqasumm} uses the chosen best answer as the answer summary, they also apply heuristics to ensure token overlap with the remaining answers. However, we found that the heuristics applied generally promotes only long answers instead of multiple perspectives. To validate our hypothesis, we examine a set of 30 summaries from CQASumm and found that only 37\% of the examples contained multi-perspective answers. 
\par
Although multi-perspective answer summarization is an important research topic with practical applications, there are no relevant datasets or techniques to address it effectively. This paper tries to close this gap by developing a dataset together with several modeling techniques for multi-perspective answer summarization. To generate a multi-perspective summarization dataset, we devise a pipeline to produce \textit{bullet point answer summaries}. First, we select and cluster salient answer sentences. Then, we use the cluster centroids as our summary bullet points and remove them from the input to promote a more challenging, more abstractive task. We further filter the data to improve our dataset's quality and promote desirable summary characteristics such as compression. We find that a strong baseline model trained on our data inherently outputs multi-perspective summaries. We focus our modeling efforts on generating content implied by the input text and being faithful to the underlying answers by covering multiple perspectives. To this end, we use a reinforcement learning (RL) framework with new rewards and a sentence-relevance multi-task loss, whereby the model learns to predict relevant sentences for the current decoding step to more closely align the source and generated output. Our models improve the coverage and faithfulness of generated summaries when compared to a state-of-the-art abstractive baseline. 
\par
The main contribution of this paper is to develop, for the first time, a method for multi-perspective abstractive answer summarization. To achieve this, 1) We introduce a dataset generation pipeline for answer summarization that goes beyond the best-answer summary, to create multi-perspective, bullet-point summaries for training and evaluation 2) We introduce and evaluate RL reward functions on answer summarization, including entailment as a measure of faithfulness and volume of semantic space as a way to increase coverage of multiple answer perspectives 3) We introduce a sentence-relevance prediction loss to increase the faithfulness and interpretability of the generated answer summaries. We will make our code available for reproducing our dataset pipeline and model results.
\section{Related Work}\label{sec:related_work}
\par \noindent
\textbf{Extractive Answer Summarization:}
Much work has focused on the extractive summarization setting as an answer-ranking problem \citep{chan-etal-2012-community,pande2013summarizing, wang-etal-2014-query}.
\citet{liu-etal-2008-understanding} find that only 48\% of the best answers on Yahoo! Answers are unique best answers; there are multiple correct ways to answer a question. Other recent work has focused on sentence extraction using metadata \citep{tomasoni-huang-2010-metadata} or sparse-coding frameworks \cite{song2017summarizing}. Our focus is on representing multiple perspectives in an abstractive summarization framework.
\par \noindent
\textbf{Abstractive Answer Summarization:}
Another line of work has attempted abstractive answer summarization by treating the tagged best answer as the gold summary of all the other answers \citep{chowdhury2019cqasumm, chowdhury2020neural}. \citet{chowdhury2019cqasumm} present CQASumm, a dataset of about 100k examples consisting of the best answer as the gold summary, which, however, often only contains one perspective. 
\par \noindent
\textbf{Multi-document Summarization:}
Answer summarization can be viewed as a query-based multi-document summarization (MDS) problem. Several large-scale MDS datasets have been introduced in the news domain \citep{fabbri-etal-2019-multi, gu2020generating, ghalandari2020large}, for creating Wikipedia lead-paragraphs \citep{liu2019generating} and for long-form question answering \citep{fan-etal-2019-eli5}. However, news-based MDS datasets are not query-based, Wikipedia summarization is topic-based and less granular than our setting, and the ELI5 dataset \citep{fan-etal-2019-eli5} summarizes web documents rather than direct query answers.
\par \noindent
\textbf{RL and Multi-task Learning for Summarization:}
\citet{paulus2017deep} first apply the REINFORCE algorithm \cite{Williams:92} in the context of summarization. RL has since been applied for both extractive \cite{narayan-etal-2018-ranking, dong-etal-2018-banditsum}, abstractive \cite{pasunuru-bansal-2018-multi, li2018ensure, huang-etal-2020-knowledge, laban-etal-2020-summary} and hybrid approaches \cite{chen-bansal-2018-fast}. \citet{bohm-etal-2019-better} stress the role of using rewards that correlate well with human judgments on downstream performance. Our paper focuses on the selection of rewards applicable for promoting faithful and diverse, abstractive answer summaries. Previous work on entailment as an RL reward has focused on document-level entailment in the news domain \citep{li2018ensure, pasunuru-bansal-2018-multi}. In this work, we show the effect of the choice of entailment model on downstream faithfulness prediction and the importance of using sentence-level entailment. 
Recent work in multi-task learning with summarization consists of sharing parameters between an abstractive generator and auxiliary tasks such as entailment and question generation \citep{guo-etal-2018-soft} and text classification and syntax-labeling tasks \citep{lu2019multi}. 
\section{Dataset Creation}\label{sec:dataset_pipeline}
Previous CQA work lacks multi-perspective supervision. To address this research gap, we develop a system to create summaries covering multiple perspectives of answers to a given question.
\PipelineFigure
\subsection{Overview of Data Generation Pipeline}
The input to our pipeline is a question and its answers. We use question threads from the Yahoo! Answers L6 corpus\footnote{\url{https://webscope.sandbox.yahoo.com/catalog.php?datatype=l&did=11}}. Our pipeline operates on the sentence level of these answers versus the answer level, as we believe that this granularity allows us to capture additional perspectives.
Our dataset pipeline consists of the following components: 1) a relevance model to remove irrelevant inputs, 2) a clustering model to cluster similar content, and 3) input and summary creation from centroids. 
\par \noindent
\textbf{Relevance model:}
We first aim to determine whether a given sentence is relevant to answering a question and, therefore, to be considered as a potential summary sentence. We use the ANTIQUE \citep{hashemi2020antique} relevance data for training a query-sentence relevance model. The data consists of Yahoo! answers and relevance labels on a scale from 1-4, with 1-2 not relevant and 3-4 relevant. We use a RoBERTa-based \cite{liu2019roberta} model fine-tuned on answer selection on the TREC-QA dataset \citep{wang-etal-2007-jeopardy} as a binary relevant/non-relevant classifier and further fine-tune it using the Tanda \citep{garg2019tanda} method. We experimented with training the relevance classifier using Yahoo! Answers, treating the best answer as relevant and the other answers as not relevant, and analogously on the sentence level, although without improvements. The performance was measured using mean reciprocal rank on a held-out relevance set. 
\par
As input to the clustering stage, we remove sentences that our relevance model labels as irrelevant (our model tends to over-predict relevant sentences, as many answers contain relevant sentences, thus removing only 16\% of sentences). Improving this relevance classifier to better filter irrelevant answer sentences is a very interesting research direction, although we leave this for future work. 
\par \noindent
\textbf{Clustering:}
Most methods for short-text clustering \citep{hadifar-etal-2019-self, xu2017self} require a known value of k, the number of clusters, which is dynamic from question to question in our setting. In this work, we use the sentence-transformers library \citep{reimers-2019-sentence-bert} to perform clustering. Specifically, we start with a RoBERTa-based model fine-tuned for sentence embeddings on an entailment dataset, which is further fine-tuned for semantic similarity. Clustering parameters were chosen based on a StackOverflow clustering dataset containing labeled clusters commonly used in short-text clustering. We used Agglomerative clustering with average linkage, cosine distance, and a maximum distance of .65.
\par
To create the final summaries, we locate the centroid of clusters with at least two sentences and use these centroids as bullet-point summaries. Further, we remove the centroid sentences from the sentence-tokenized input answers to create a challenging abstractive summarization dataset analogous to the XSum dataset \citep{narayan-etal-2018-dont}. Since each cluster contains at least two sentences, we assume that given a perfect clustering algorithm, a related sentence can help generate the removed centroid sentence. While removing sentences naturally decreases coherence, we believe that this introduces a tolerable level of noise, considering the existing presence of noise through ungrammatical and occasionally incoherent answers. To further account for imperfections in the pipeline, we apply additional filtering techniques, described below.
\subsection{Postprocessing and Quantitative Analysis}
We obtained question threads from Yahoo! Answers and applied heuristics detailed in \citet{tomasoni-huang-2010-metadata} to find threads suitable for summarization. Threads were removed if 1) there were less than five answers, 2) the longest answer was over 400 words, 3) the sum of the length of all answers was outside of (100, 1000) words, and 4) the average length of answers was outside of the (50, 300) words interval. This filtering left us with about 350k of the approximately 4.4 million threads and included both factoid and non-factoid questions. Questions include the subject of the post as well as the content of the post when available. 
\par \noindent
\textbf{Example Filtering:}
We remove examples from the dataset based on desired summarization characteristics.  A desirable trait in summarization datasets is compression, i.e., the ratio of the input size to the summary size \citep{grusky-etal-2018-newsroom}. We remove examples with a compression ratio under 4, examples for which the input length exceeded 1,100 tokens and for which the summary length exceeded 250 tokens, leaving us with 284,295 examples.  We further remove target summaries labeled as contradictions from a RoBERTa-based entailment model following \citet{matsumaru-etal-2020-improving}. Furthermore, we remove examples with more than 10 “+” or "=" signs (math queries), those with very long ($>$50 characters) tokens, and those with a link in the target or more than one link in the source. Finally, we filter to ensure that we have examples where the named entities found in the target are also found in the source document. 
\par \noindent
\textbf{Quality Analysis:} The filtering process yielded 96,701 examples, which we split into 88,512/4,032/4,157 training, validation, and testing examples. We annotated a subset of 400 summaries created by our pipeline to conduct quality checks.
For each summary, the annotator reads the question, and if the answer coverage of the summary was determined as reasonable, the summary was marked as 1, otherwise 0. 370 of the 400 summaries were labeled as 1, showing that the pipeline creates largely relevant content. Additionally, on examining 30 summaries, we found that 80\% contained multiple perspectives versus the 37\% we found in CQASumm, showing the benefit of our dataset pipeline in encoding multiple viewpoints. To further analyze the types of questions present in our dataset, we trained a factoid/non-factoid question classifier using SQuAD \cite{rajpurkar-etal-2016-squad} data as factoid examples and non-factoid Yahoo! Questions dataset\footnote{\url{https://ciir.cs.umass.edu/downloads/nfL6/}} as non-factoid examples. 8\% of threads were labeled as factoid questions; the filtering steps based on answer size likely filter out examples with short, factoid answers. 
\subsection{Relation to Existing Datasets}
CQASumm is the closest dataset with our desired answer summarization qualities, although it simply promotes answers as summaries rather than truly summarizing answers. As discussed above, this dataset lacks our desired multi-perspective summaries. A similar approach to dataset creation was taken by \citet{shapira2020massive} for review summarization by clustering reviews using pivot clustering, adding reviews to a cluster based on lexical overlap until a max length and min number of review requirements are met. There are significant differences to our approach in terms of granularity (reviews vs. sentence clustering), type of clustering (lexical vs. embedding-based), as well as the ultimate use of these clusters (they train a cluster summarizer while we combine cluster centroids for creating an abstractive bullet point combined with other cluster centroids).  We present a comparison of dataset statistics between our dataset, which we call \textbf{AnswerSumm}, and the standard XSum and CNN-Daily Mail \cite{nallapati-etal-2016-abstractive} summarization datasets in Table \ref{tab:statistics}. In general, we find our dataset to be more abstractive than CNN-DailyMail and less so than XSum. We also note that our generated dataset is similar to CNN-DailyMail in that it consists of bullet points. While this may create summaries with less coherence, or potentially contradictory answers, we focus on producing multi-perspective summaries in this work and leave improved summary coherence for future work. 
\DatasetComparison
\section{RL Rewards and Auxiliary Losses}\label{sec:dataset_pipeline}
Cross-entropy loss suffers from exposure bias and also does not directly optimize the evaluation metrics such as NLI and ROUGE-L \cite{ranzato2015sequence}. The REINFORCE algorithm \cite{Williams:92}, on the other hand, allows for optimizing the evaluation metrics using non-differentiable rewards. Therefore, we use an RL multi-reward objective to promote summaries with both high coverage of the input answers and faithfulness. Additionally, we also introduce an auxiliary loss function for more interpretable and faithful summaries. 
\subsection{Multi-Reward Optimization}
We follow the settings of \citet{pasunuru-bansal-2018-multi} for optimizing multiple rewards. In the equations which follow, $x = \{x_1,\: x_2,\: \dots,\: x_{n'}\}$ refers to the input source tokens (e.g. a question and its answers), and $y^{*} = \{y^{*}_1,\: y^{*}_2,\: \dots,\: y^{*}_{N}\}$ refers to the gold target summary which consists of $\{y^{*}_{1_s},\: y^{*}_{s_s},\: \dots,\: y^{*}_{N_s}\}$ sentences.  Standard training minimizes the negative log-likelihood (NLL) loss using teacher forcing \cite{williams1989learning}:
\begin{equation}
      L_{ml} =  -\sum_{t=1}^{N} \log p(y^{*}_{t}|y^{*}_1,...,y^{*}_{t-1}, x)
\end{equation}
For our RL optimization, we use self-critical policy gradient training as in \citet{paulus2017deep, rennie2017self}. At each time-step, we produce an output $y^{s}$ by sampling from the current decoding probability, $p(y^{s}_{t}|y^{s}_1,...,y^{s}_{t-1}, x)$, as well as an output $\hat{y}$ obtained by greedily decoding from the current probability distribution. We define a reward function $r(y, x, y^{*}) \in [0,1]$, i.e., the reward function compares $y$ with $x$ and $y^{*}$. The RL loss function $L_{rl}(x,y^*)=$: 
\begin{equation}
       (r(\hat{y}, x, y^*) - r(y^{s}, x, y^*))  \sum_{t=1}^{N} \log p(y^{s}_{t}|y^{s}_1,...,y^{s}_{t-1}, x)
\end{equation}
As in \citet{paulus2017deep} and \citet{pasunuru-bansal-2018-multi}, we use a mixture of the above two losses:

\begin{equation}
     L_{mixed}  = \gamma_{rl} L_{rl} + \gamma_{ml} L_{ml}, 
\end{equation}
where $ \gamma_{rl}$ and $\gamma_{ml}$ are tunable hyperparameters used as scaling factors. Rather than applying weights to each reward, we follow \citet{pasunuru-bansal-2018-multi} and optimize $L_{mixed}$ by alternating rewards in each minibatch. 
\subsection{Rewards}
We now describe the three RL reward functions used: (1) ROUGE \cite{lin-2004-rouge} as a proxy for content coverage, (2) entailment (NLI) for faithfulness, and (3) semantic area to measure the coverage of a summary in a semantic space. 
\par \noindent
\textbf{ROUGE} \cite{lin-2004-rouge}: Similar to \citet{paulus2017deep} and \citet{pasunuru-bansal-2018-multi}, we use ROUGE-L as a reward to additionally promote important content beyond the cross-entropy loss. 
\NLIMotivation
\par \noindent
\textbf{Natural Language Inference (NLI) for Faithful Summarization}: We use the degree of entailment of summaries given input answers as a reward to promote faithfulness of answer summarization. While entailment has been used as a reward as well as a summarization metric, we find several gaps in the current literature. Firstly, a discussion of the effect of the quality of the NLI evaluation model on downstream faithfulness metrics is incomplete. Also, summarization work typically uses NLI models with document-level input, while NLI models are generally trained on sentence-level data.
\par
\citet{falke-etal-2019-ranking} analyze NLI models for ranking summaries; given an input sentence and two summary sentences, one faithful and one unfaithful to the input, a model should rank the faithful summary higher than an unfaithful summary. They introduce a dataset of 377 examples and measure the rank accuracy of NLI models. They define NLI as a measure of faithfulness for ranking summaries in the following way: Let $\mathcal{N}$ be an NLI model which, given a claim $c$ and a premise $p$, computes $\mathcal{N}(p, c)$, the probability that the claim is entailed by the premise. We use this to calculate the NLI score for a summary $y$ consisting of $N_s$ sentences:
\begin{equation}
      \text{NLI}(y, x) =  \frac{1}{N_{s}} \sum_{i=1}^{N_{S}}  \max_{s \in x} \mathcal{N}(s, y_{i_s}) 
\label{eq:nli}
\end{equation}
\par
For the original task introduced in \citet{falke-etal-2019-ranking}, $x$ consists of a single source sentence from the CNN-DailyMail corpus. We present our findings on this task in Table \ref{tab:nli_motivation}. We examine how the quality of the NLI model affects performance by comparing BART \cite{lewis-etal-2020-bart} and RoBERTa fine-tuned on the MNLI corpus \cite{williams-etal-2018-broad}. Although the performance gap of these two models is very small on MNLI (90.2\% for RoBERTa and 89.9\% for BART), the performance gap is very large on ranking these sentences (89.8\% for RoBERTa and 71.9\% for BART).
\par
We also address the effect of the granularity of the NLI model input. As discussed above, \citet{falke-etal-2019-ranking} perform ranking based on sentence-level input and output. Recent work in entailment as a summarization metric, however, uses the entire input document as input to the NLI model for faithfulness calculations \citep{maynez-etal-2020-faithfulness}, rather than computing the max over all the input sentences as in Equation \eqref{eq:nli}. We locate the full source articles for the 377 examples and perform two experiments, one using Equation \eqref{eq:nli}, and the other which uses the entire article to score the target sentence, $\mathcal{N}(x, y_{i_s})$. Performance drops when using the entire article as the input versus using Equation \eqref{eq:nli}. To ensure that the performance drop was not caused by content truncation due to the 512 input size limitation, we also experimented with using the article starting from the relevant source sentence, without improvements.
\par
Furthermore, we find that the use of NLI is particularly suitable for AnswerSumm. We sampled six threads from our dataset. Then for each thread, we wrote sentences entailed by the source as well as sentences based on similar themes but not stated in the source, totaling 50 faithful and 50 hallucinated examples. 
We find that the RoBERTa MNLI model can correctly identify these examples with 96\% accuracy. We believe that NLI is intuitively more suitable for our data, which is less entity-heavy when compared to the news domain.
\par \noindent
\textbf{Semantic Area for Multi-Perspective Summarization}: We aim to reward summaries that include more of the perspectives found in the input answers. To achieve diverse extractive summarization, \citet{yogatama-etal-2015-extractive} embed sentences in semantic space and select those whose convex hull maximizes the volume in that space.
This idea of semantic volume is also used to measure the semantic overlap between summaries and references in \citet{jung-etal-2019-earlier}. We use semantic volume as a proxy for covering multiple perspectives; the summary with the larger semantic volume covers a wider range of views discussed in the input. We make use of sentence-transformers \cite{reimers-gurevych-2019-sentence} to obtain sentence embeddings for each sentence. We project each embedding onto two dimensions using PCA, and thus, our volume calculation reduces to an area calculation, which we call \textbf{Semantic Area}. We use min-max normalization to keep the reward in the range of 0 to 1. 
\subsection{Relevant Sentence Prediction}
We want to more closely align the decoded summary with the source text, as hallucinations may be caused by the decoder acting more as a language model rather than attending to the source text \cite{maynez-etal-2020-faithfulness}. Aligning the source and generated output offers a potential interpretable output during inference, which goes beyond using attention for interpretation \cite{wiegreffe-pinter-2019-attention}. We introduce an auxiliary loss by which the model predicts, based on the decoder representation, a span of source text relevant to the current time-step, analogous to finding evidence to support a claim of factuality \cite{kryscinskiFactCC2019}. 
\par
Let $h_{e_i}\in R^{dim_e}$ be the representation of token $x_i$ from the last layer of the encoder. Let
$h_{d_i}\in R^{dim_d}$ be the representation of token $y^{*}_i$ from the last layer of the decoder right before the softmax layer. Here, $dim_{e} = dim_{d} = 1024$. Let $h_{e}$ be the concatenation of all $h_{e_i}$ and $h_{d}$ be the concatenation of all $h_{d_i}$. We then pass these representations through separate layers $L_e$ and $L_d$ which correspond to the typical layer used in BART classification tasks except that it outputs a representation of size 2048:
\begin{equation}
    h^{*}_{e} = L_e(h_{e}), \; h^{*}_{d} = L_d(h_{d})
\end{equation}
We split the resulting representations in half along the hidden dimension, resulting in encoder representations $h^{*}_{e{\text -}start}, h^{*}_{e{\text -}end}$ and decoder representations $h^{*}_{d{\text -}start}, h^{*}_{d{\text -}end}$ which will be used for start and end relevant source span prediction. We then compute an inner product between these representations, resulting in logits over the input corresponding to potential start and end spans: 
\begin{equation}
\begin{split}
    \text{logit}_{start} = h^{*}_{e{\text -}start} \bullet h^{*T}_{d{\text -}start} \\
     \text{logit}_{end} = h^{*}_{e{\text -}end} \bullet h^{*T}_{d{\text -}end}
    \end{split}
\end{equation}
Cross entropy loss can then be calculated over the start and end logits with reference to gold spans as in SQuAD question answering training. We call this loss $L_{span}$. Our final loss function becomes: 
\begin{equation}
     L_{mixed}  = \gamma_{rl} L_{rl} + \gamma_{ml} L_{ml} + \gamma_{span} L_{span}, 
\end{equation} 
where $L_{span}$ is the cross-entropy loss over start and end span predictions. Specifically, we separate input sentences with special tokens and predict sentence-level spans, which amounts to predicting a start and end token corresponding to a relevant sentence, so we call this model variation \textbf{Sent Relevance}. For each sentence in the gold target training data, we calculate the BM25 scores of the sentences in the source to pick the gold relevant source sentence for that target sentence. All the timesteps corresponding to a target sentence use the same relevant input sentence. We also experimented with just predicting relevant source sentences at the end of each target, using a binary sentence classification loss and a regression loss over the BM25 scores, without significant improvements.
\BaselineResults
\section{Experimental Settings}\label{sec:experimental_settings}
We use the fairseq codebase \cite{ott-etal-2019-fairseq} for our experiments. Our base abstractive text summarization model is BART \cite{lewis-etal-2020-bart}, a pretrained denoising autoencoder that builds off of the sequence-to-sequence transformer of \citet{vaswani2017attention}. Input to the model is the question concatenated with input answers.
We fine-tune BART using a polynomial decay learning rate scheduler with learning rate $3\mathrm{e}{-5}$, using the Adam optimizer \citep{kingma2014adam}. We train with 500 warmup steps and 20,000 total steps and pick the model with the best label-smoothed cross-entropy \cite{szegedy2016rethinking} validation loss. Cross-entropy loss is our main loss, while the RL rewards and sentence-relevance prediction can be viewed as auxiliary losses. In RL experiments, we train using BART from scratch, as opposed to using a model already fine-tuned on answer summarization, as we found that this model better learned to follow the given rewards. Following similar ratios as in \citet{lu2019multi}, we set ($\gamma_{rl}$,$\gamma_{ml}$,$\gamma_{span}$) =  (0.9, 0.1, 0.0) when experimenting without sentence-relevance loss, (0.00, 1.0, 1.0) for experiments with just relevant sentence prediction and cross-entropy loss, and (0.9, 0.5, 0.01) for experiments with all losses. Hyperparameters were tuned on the validation set; we found a larger $\gamma_{ml}$ necessary when combining rewards with sentence relevance prediction to ensure that the main negative log-likelihood loss was not drowned out by the auxiliary losses. 
\section{Results}\label{sec:results}
\par \noindent
\textbf{Extractive Baselines}: We use standard extractive summarization baselines such as Lexrank \cite{erkan2004lexrank} and TextRank \cite{mihalcea2004textrank}, and a BERT-based extractive model, BertSum \cite{liu2019text}. Results are presented in Table  \ref{tab:baseline_results}. We observe a large gap between these baselines and the extractive oracle, which is the upper bound for extractive model performance, showing potential for improvement. Since we focus on abstractive summarization, we leave improving extractive models for future work.
\MainResults
\par \noindent 
\textbf{Abstractive Models:}
We present the results of the abstractive models in Table \ref{tab:main_results}. We note that while the model with ROUGE reward outperforms the baseline in ROUGE-L (the ROUGE variant optimized), we do not see larger gains in ROUGE due to the similarity between the ROUGE optimization and NLL on our datasets. For bullet-point summaries, minimizing the NLL is analogous to rephrasing relevant bullet-points from the source and increasing the ROUGE-L. The model that combines all the RL rewards achieves the highest ROUGE performance, while the model with all RL rewards and sentence-relevance loss achieves the highest NLI score. The faithfulness of the model with only sentence relevance loss is further improved by adding the RL rewards. In general, we see that the model with a single RL reward achieves the highest score of the target summaries for that metric, i.e., the highest NLI score is achieved using only the NLI-based reward. Additionally, we calculate the average semantic area of the resulting summaries. The baseline model, the model with just semantic reward, and the final model with all rewards have semantic areas of 39.7, 46.5, and 42. To further show the effect of our dataset on multi-perspective summarization, we train a BART model on the most related answer summarization dataset CQASumm and find that the semantic area of that model's summaries is 31.54. This result shows the importance of supervision from our dataset pipeline for ensuring coverage of multiple perspectives in answer summarization. 
\par
As automatic metrics may not correlate perfectly with human judgments, we perform a human evaluation to determine the differences in model output qualities. We presented two annotators who are fluent in English with the question, answers, and summaries from three models and asked them to annotate the summaries along the following dimensions: 1)  On a Likert scale from 1-5, label the ability to capture multiple perspectives, with points deducted for repetition 2)  On a Likert scale from 1-5, label the extent of faithfulness of the summary, with 5 being a completely faithful summary and 1 being an entirely inaccurate summary. 
\par
We present results in Table \ref{tab:human_evaluation}. Annotations are averaged between each annotator and then across examples for 50 questions threads from three models. We choose to compare the BART baseline, the BART model with all RL rewards, and the BART model with span prediction to determine the effect of our rewards and the multi-task loss on output quality. Pearson correlations for faithfulness and multi-perspective scores among the annotators were 0.41 and 0.31, displaying moderate correlation. We find that most models can generate multiple perspective summaries. The baseline already generates multi-perspective output, likely because the dataset pipeline produces summaries that contain multiple perspectives, so the baseline learns to produce such output. Using a student's t-test with a p-value of 0.05, we find that the improvement in faithfulness between the RL models and the baseline is statistically significant while the other differences are not. With the span-based model, this improvement comes at the cost of some level of abstraction, as the percentage of novel unigrams found in the summary is 10\% vs. 13\% found in the baseline and RL-only models. This reduction in abstraction likely results because the span loss more closely binds the decoder representation with the encoder representation, encouraging the model to copy more from the source. We demonstrate the added advantage of our span prediction model's interpretability by using it to provide explanations for the generated summaries in the Appendix.
\HumanAnnotation
\section{Conclusion and Future Work}\label{sec:conclusion}
We propose multi-perspective answer summarization by introducing a pipeline for creating a suitable dataset for the task and by introducing models to promote high-coverage, faithful answer summaries, as seen in automatic and human evaluations. We aim to refine this pipeline for future work by improving the relevance and clustering components and applying them to new data sources. We plan to study the abstractiveness-faithfulness tradeoff further, explore additional rewards for improved summary coherence, and move beyond bullet point summaries by building on work in sentence fusion. 
\section{Ethical Considerations}\label{sec:ethical_considerations}
As we propose a novel conversation summarization dataset creation pipeline and modeling components, this section is divided into the following two parts.
\subsection{New Dataset}
\paragraph{Intellectual Properties and Privacy Rights}
We will be providing scripts to run our dataset creation pipeline but will not be releasing the data itself. Access to the Yahoo! Answers dataset requires the submission of a user-agreement form through the webscope platform \footnote{\url{https://webscope.sandbox.yahoo.com/catalog.php?datatype=l&did=11}}. We do not do any crowdsourcing. All human annotations are done in-house. We manually reviewed our dataset output for quality and potential problems. 
\subsection{NLP Application}
\paragraph{Bias}
Biases may exist in the datasets, such as political bias and gender bias in Yahoo! Answers. Thus, models trained on these datasets may propagate these biases.
\paragraph{Misuse Potential and Failure Mode}
When used as intended, applying the summarization models described in this paper can save people much time. However, the current models are still prone to producing hallucinated summaries, and in such a case, they may contribute to misinformation on the internet. We move the needle in faithful summarization in this paper, but further research is needed to ensure the faithfulness of abstractive summaries to address this issue, as this issue is present among all current abstractive summarization models. 
\paragraph{Environmental Cost}
The experiments described in the paper make use of V100 GPUs. We used up to 8 GPUs per experiment. The experiments may take several hours. Several dozen experiments were run due to parameter search, and future work should experiment with distilled models for more light-weight training. We note that while our work required extensive experiments to draw sound conclusions, future work will be able to draw on these insights and need not run as many large-scale comparisons. Models in production may be trained once for use using the most promising settings.  

\section{Appendix}\label{sec:appendix}
We show the model-generated summaries for the model "BART + Sent Relevance + RL (ALL)" as well as the sources sentences the model predicts as relevant at the end of each sentence generated during decoding.  In the example below, the model can correctly abstract meaning from the source sentences and formulate summary bullet points. Occasionally the model will output a point which is not coherent by itself (e.g. 'It's a great book') or may output related but not supported text. We believe this is due to the BM25 relevance function used for determining relevant sentences for training. Examining this mechanism sentence relevance prediction as a model probing task as well as improving coherence in summaries, to go beyond bullet point summaries via methods such as sentence fusion, are valuable areas of future work. 
\par \noindent
We show model outputs from the three models examined in human evaluation in Table \ref{tab:example_summaries_1}. We see that the baseline hallucinates several times. We also notice how the hallucinations, as opposed to typical hallucinations in the news domain which may replace entities, are often plausible responses. For example, although the baseline generates an output saying that it is not a good idea to lose weight, which is not directly stated in the source, such an answer is very plausible. We also found that there was occasionally a fine line between what was a hallucination and what was a plausible generated text which is not entirely implied in the source. For example, the text stating it is not a good idea to lose weight echoes the sentiment that the user asking the question should make the choice for themselves, although this is not stated in this fashion. We believe that more precisely defining the degrees of hallucination and plausibility to be an important direction for future work. 
\clearpage
\SpanPredictionExample
\ExampleSummaries

\bibliographystyle{acl_natbib}
\bibliography{acl2021}


\end{document}